\documentclass[10pt,letterpaper,twocolumn]{article}

\usepackage{newtxtext,newtxmath}
\usepackage[left=0.5in,right=0.5in,bottom=0.5in,top=1in]{geometry}
\usepackage[round,authoryear]{natbib}

\usepackage{titlesec}
\usepackage{microtype}
\usepackage{booktabs}
\usepackage{makecell}
\usepackage{array}
\usepackage{balance}

\titleformat{\section}{\bfseries\centering}{}{0pt}{\MakeUppercase}
\titlespacing*{\section}{0pt}{0.70\baselineskip}{0.30\baselineskip}

\titleformat{\paragraph}[runin]{\itshape}{}{0pt}{}[.\ ]
\titlespacing{\paragraph}{0.25in}{0pt}{0.5em}

\begin{document}

\twocolumn[{%
\begin{center}
{\bfseries\large Interpretable Modeling of Driver Attention Shifts\\
with a Vision--Language Model}\\[0.65em]

Kaiser Hamid$^{1}$, Khandakar Ashrafi Akbar$^{2}$, Peihang Li$^{1}$, and Nade Liang$^{1}$\\
$^{1}$Texas Tech University, Lubbock, Texas, USA\\
$^{2}$Towson University, Towson, Maryland, USA\\[0.45em]

\end{center}
}]

\begin{abstract}
Driver gaze is commonly modeled as a spatial heatmap, but heatmaps
alone are difficult for humans to interpret because they do not explain
which road object or region is being monitored or why an attention shift
may matter. This study examines whether minimal human-grounded
supervision can steer a vision--language model toward interpretable
descriptions of driver attention shifts. Using selected high-change
gaze moments from the Berkeley DeepDrive-Attention dataset, we compare
zero-shot, one-shot, and LoRA fine-tuned VLM conditions against
human-refined reference descriptions and expert ratings. Results show
that fine-tuning with 80 expert-refined attention examples improves
ROUGE-L, METEOR, Entity Alignment F1, and Human Alignment Score relative
to unsteered VLM outputs. The findings suggest that language-based
descriptions can complement gaze heatmaps by making driver attention
more accessible for human-factors analysis, driver-monitoring review,
and situation-awareness support.
\end{abstract}

\section*{INTRODUCTION}

Driving is a highly visual task. Where a driver looks---and when
attention shifts---directly shapes hazard detection, decision-making,
and crash risk \citep{fang2022dada}. Advanced driver-assistance systems
that can anticipate and explain a driver's attentional focus could
support operator monitoring, timely alerts, and smoother human--AI
handoffs in partially automated vehicles.

Current driver-attention models often represent gaze as a spatial
heatmap. Although useful for prediction, a heatmap alone is difficult
to interpret: it does not specify which road user, signal, lane region,
or potential hazard is being monitored, nor why attention shifted at a
particular moment \citep{alletto2016dreyeve,xia2019bdda}. This limits
the value of gaze modeling for human-centered driver-monitoring
interfaces, where system outputs must be understandable to designers,
operators, and safety analysts.

This study tests whether minimal human-grounded supervision can steer a
general-purpose vision--language model (VLM) toward descriptions of
driver attention shifts that align with expert interpretation of where
the driver is looking. The goal is not to replace gaze maps, but to
make gaze shifts more interpretable by linking attended locations to
road objects, scene context, and situation-awareness-relevant meaning.

\section*{BACKGROUND}

Driving-attention research has progressed from saliency-based visual
attention modeling toward more semantic accounts of what drivers attend
to in complex scenes \citep{borji2013attention,palazzi2019dreyeve}.
Language-based driving research further suggests that naming
risk-relevant objects and events can support more interpretable accounts
of driver behavior than spatial maps alone
\citep{mori2019attention,malla2023drama,zhou2025where}. Recent VLMs
provide a promising interface for this goal because they connect visual
evidence with natural-language descriptions \citep{liu2023llava,
tian2024drivevlm}. However, generic VLM scene descriptions are not
automatically aligned with human gaze behavior in safety-critical
driving.

Existing VLM-based approaches often emphasize scene-level reasoning or
spatial gaze prediction and still require substantial supervision. This
work asks a narrower human-factors question: can a small amount of
human-refined supervision steer VLM outputs toward descriptions that a
domain expert would independently produce when interpreting driver
attention? This framing keeps the emphasis on interpretability, expert
judgment, and driver-monitoring use rather than on model architecture
alone.

\section*{APPROACH}

\paragraph{Attention-shift sampling}
We use the Berkeley DeepDrive-Attention (BDD-A) dataset, which contains
dashcam video frames paired with driver gaze annotations
\citep{xia2019bdda}. Rather than sampling all frames uniformly, we
focus on moments where attention changes substantially. Consecutive
gaze maps are treated as spatial distributions, and high-change frame
pairs are retained as attention-shift instances. This produced 80
training, 20 validation, and 81 test samples.

\paragraph{Human-grounded descriptions}
For each selected attention-shift moment, GPT-4o drafts a concise
description of the scene, the attended region, the likely next gaze
target, and why that target may matter. Domain experts then review the
drafts for factual accuracy, object specificity, and consistency with
the gaze evidence. This human-in-the-loop step moves the supervision
away from generic scene captioning and toward descriptions grounded in
driver attention.

\paragraph{VLM comparison}
The comparison separates three effects: general VLM capability, response
guidance, and human-grounded attention supervision. Zero-shot LLaVA
\citep{liu2023llava} tests whether a general-purpose VLM can describe
driver attention without task-specific steering. One-shot LLaVA tests
whether limited examples can impose the desired response pattern. The
fine-tuned model tests whether 80 human-refined attention examples
improve alignment with expert interpretation using parameter-efficient
LoRA adaptation \citep{hu2022lora}.

\begin{table}[t]
\centering
\footnotesize
\setlength{\tabcolsep}{4pt}
\renewcommand{\arraystretch}{0.95}

\begin{tabular}{@{}lcccc@{}}
\toprule
Setup & ROUGE-L & METEOR & \makecell{Entity\\Alignment F1} &
\makecell{Human\\Alignment Score} \\
\midrule
Zero-shot LLaVA  & 0.171 & 0.302 & 0.18 & 1.25 \\
One-shot LLaVA   & 0.336 & 0.397 & 0.29 & 2.83 \\
Fine-tuned model & \textbf{0.573} & \textbf{0.581} &
\textbf{0.51} & \textbf{3.27} \\
\bottomrule
\end{tabular}

\vspace{5pt}
\refstepcounter{table}\label{tab:main}
{\raggedright\footnotesize
\textbf{Table \thetable.} Held-out evaluation of VLM conditions. Human
Alignment Score ranges from 1 = very poor alignment to 5 = very good
alignment.\par}
\vspace{-4pt}
\end{table}

\section*{OUTCOME}

Outputs were evaluated against human-refined reference descriptions
using ROUGE-L, METEOR, Entity Alignment F1, and Human Alignment Score.
ROUGE-L and METEOR assess overlap with the reference descriptions.
Entity Alignment F1 is the most directly interpretable automatic
measure for this task because it asks whether the model identifies the
same road object or region as the human-refined reference, such as a
lead vehicle, pedestrian, lane region, or traffic signal.

For the expert evaluation, generated descriptions were first produced
for the held-out test set under each model condition. Two domain
experts then evaluated 30 randomly sampled test cases. The model
condition was hidden from the raters, so experts did not know whether a
description came from the zero-shot, one-shot, or fine-tuned model. For
each case, experts inspected the driving scene, the gaze evidence, and
the human-refined reference description before assigning a Human
Alignment Score from 1 to 5. The rating reflected how well the generated
description matched the expected interpretation of where the driver was
looking and why that attended location was relevant.

This expert-rating procedure was included because lexical similarity
alone is not sufficient for a human-factors evaluation of attention
interpretation. A generated description can share words with the
reference while still missing the relevant road object or the reason
the gaze shift matters. Conversely, a description may use different
wording but still identify the same attended object or region. The
human evaluation therefore tested whether the output was useful as an
interpretable account of driver attention, not only whether it matched
the reference text surface form.

Table~\ref{tab:main} shows that general VLM capability alone was not
sufficient. Zero-shot LLaVA produced relatively generic scene
descriptions with low entity alignment and low human alignment.
One-shot LLaVA improved the response pattern, but the fine-tuned model
performed best across all reported metrics. Entity Alignment F1
increased to 0.51, and Human Alignment Score increased to 3.27 out of
5. This indicates moderate expert agreement and a clear improvement
over unsteered VLM outputs, but not yet deployment-ready explanation
quality.

\section*{DISCUSSION}

This study shows that driver gaze can be represented not only as a
spatial heatmap, but also as an interpretable account of the road object
or region being monitored. This matters for human factors because gaze
heatmaps are useful for computational modeling but are difficult for
humans to audit directly. A language description can make the same
attention shift easier to inspect, compare with expert judgment, and
use in driver-monitoring analysis.

The comparison among zero-shot, one-shot, and fine-tuned VLM conditions
also clarifies what type of supervision is needed. Zero-shot generation
tested whether general VLM scene understanding was enough. One-shot
generation tested whether the model mainly needed response-pattern
guidance. The fine-tuned condition tested whether human-refined
attention examples add information beyond formatting. The observed
pattern suggests that response guidance helps, but expert-refined
attention examples are needed to make the output more consistent with
human interpretation of the driving scene.

From a driver-monitoring perspective, these descriptions could support
post-event review, dataset annotation, operator-facing explanations,
and safety analysis. A heatmap may indicate that gaze moved toward the
upper center of the scene, but a language description can indicate that
the driver is likely monitoring a traffic signal, lead vehicle, or
potential conflict area. This makes the output more accessible for
human factors researchers and system designers.

The scope of this study is the human-factors representation of driver
attention: how gaze shifts can be translated into interpretable
descriptions that support expert review, driver-monitoring analysis,
and situation-awareness assessment. The current approach focuses on
language descriptions of attention shifts and does not jointly optimize
spatial gaze prediction with language generation. This creates an
important future direction: integrating spatial attention prediction
with language-based explanation in a unified driver-attention modeling
framework. The evaluation also uses a compact BDD-A subset, so the
results provide feasibility evidence rather than broad generalization.
Expert-refined descriptions improve interpretability but introduce
annotation cost and possible reviewer subjectivity.

Future work should integrate spatial attention prediction with
language-based explanation, evaluate larger naturalistic driving
datasets, and examine alignment with behavioral measures of driver
attention. Real-time deployment was not evaluated here; future work
should also benchmark latency, model compression, and streaming
inference for driver-monitoring use.

\section*{CONCLUSION}

This study demonstrates that minimal human-grounded supervision can
improve VLM descriptions of where drivers look during attention shifts.
The main contribution is a human-centered representation of gaze
behavior: making attention shifts more interpretable for driver
monitoring and situation-awareness support. The results suggest that
expert-refined language supervision is a feasible step toward more
transparent driver-attention modeling, while also identifying the need
for larger-scale evaluation and integrated spatial-language attention
models.

\section*{DECLARATION OF CONFLICTING INTERESTS AND FUNDING}

The authors declared no potential conflicts of interest. No external
funding was received for this work.


{\fontsize{8}{9}\selectfont

}

\end{document}